\documentclass[runningheads]{llncs}

\usepackage{eccv}

\usepackage{eccvabbrv}

\usepackage[normalem]{ulem}
\usepackage{float}
\usepackage{colortbl}
\usepackage{algpseudocode}
\usepackage{algorithm}
\usepackage{multirow}
\usepackage{wrapfig}
\usepackage{graphicx}
\usepackage{booktabs}
\usepackage{marvosym}
\usepackage[accsupp]{axessibility}  

\usepackage[pagebackref,breaklinks,colorlinks,citecolor=eccvblue]{hyperref}

\def\blue#1{\textcolor{blue}{#1}}

\def\eg{\emph{e.g.}}

\def\ie{\emph{i.e.}}

\definecolor{mygray}{gray}{.9}
\usepackage{orcidlink}

\begin{document}

\title{High-order Neighborhoods Know More: \\ 
HyperGraph Learning Meets Source-free Unsupervised Domain Adaptation } 

\titlerunning{Abbreviated paper title}


\author{Jinkun Jiang$^1$,
Qingxuan Lv$^1$,
Yuezun Li$^{1,\text{\Letter}}$,
Yong Du$^1$, \\
Sheng Chen$^2$,
Hui Yu$^3$,
Junyu Dong$^{1,\text{\Letter}}$}

\renewcommand{\thefootnote}{}
\footnotetext{\Letter: Corresponding author.}
\renewcommand{\thefootnote}{\arabic{footnote}}
\authorrunning{F.~Author et al.}

\institute{$^1$ Ocean University of China \quad
$^2$ University of Southampton \\
$^3$ University of Portsmouth \\
}

\maketitle

\begin{abstract}
Source-free Unsupervised Domain Adaptation (SFDA) aims to classify target samples by only accessing a pre-trained source model and unlabelled target samples. Since no source data is available, transferring the knowledge from the source domain to the target domain is challenging. Existing methods normally exploit the pair-wise relation among target samples and attempt to discover their correlations by clustering these samples based on semantic features. The drawback of these methods includes: 1) the pair-wise relation is limited to exposing the underlying correlations of two more samples, hindering the exploration of the structural information embedded in the target domain; 2) the clustering process only relies on the semantic feature, while overlooking the critical effect of domain shift, \ie, the distribution differences between the source and target domains. To address these issues, we propose a new SFDA method that exploits the high-order neighborhood relation and explicitly takes the domain shift effect into account. Specifically, we formulate the SFDA as a Hypergraph learning problem and construct hyperedges to explore the local group and context information among multiple samples. Moreover, we integrate a self-loop strategy into the constructed hypergraph to elegantly introduce the domain uncertainty of each sample. By clustering these samples based on hyperedges, both the semantic feature and domain shift effects are considered. We then describe an adaptive relation-based objective to tune the model with soft attention levels for all samples. Extensive experiments are conducted on Office-31, Office-Home, VisDA, and PointDA-10 datasets. The results demonstrate the superiority of our method over state-of-the-art counterparts. 
  \keywords{Source-free domain adaptation \and Unsupervised learning}
\end{abstract}

\vspace{-1cm}
\section{Introduction}
\label{sec:intro}

\vspace{-0.2cm}
\begin{figure}[!ht]
  \centering
  \begin{subfigure}[t]{0.37\linewidth}
    \centering
    \includegraphics[width=\linewidth]{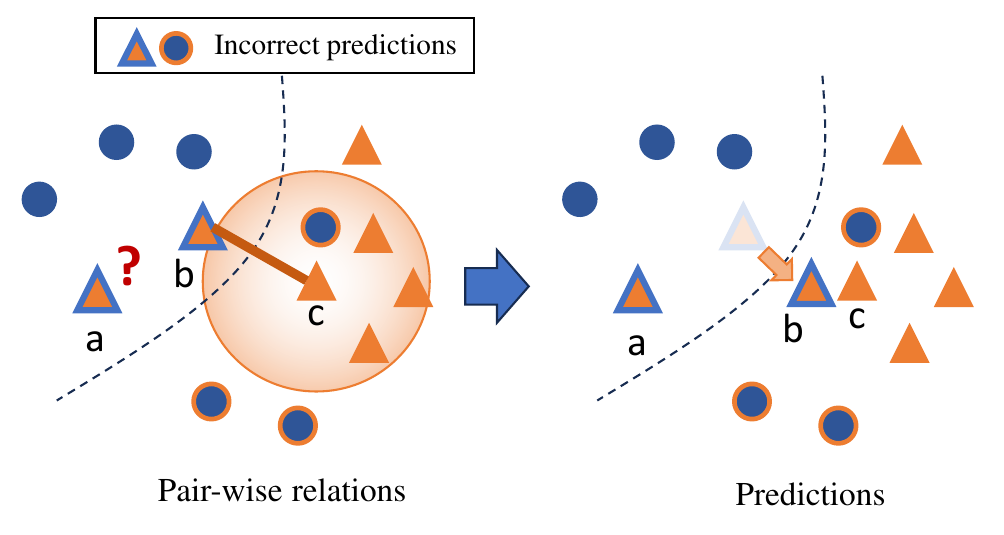}
    \label{fig:veri-c}
    
  \end{subfigure}
  \begin{subfigure}[t]{0.62\linewidth}
    \centering
    \includegraphics[width=0.45\linewidth]{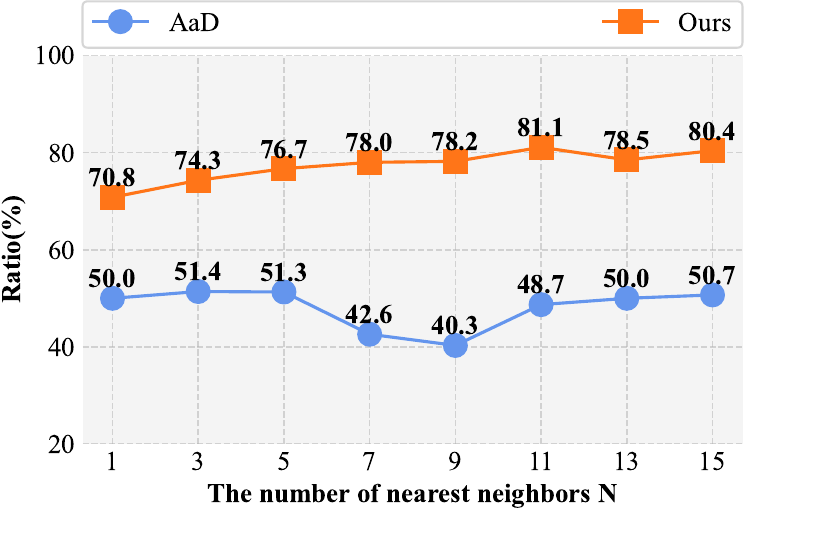}
    \includegraphics[width=0.45\linewidth]{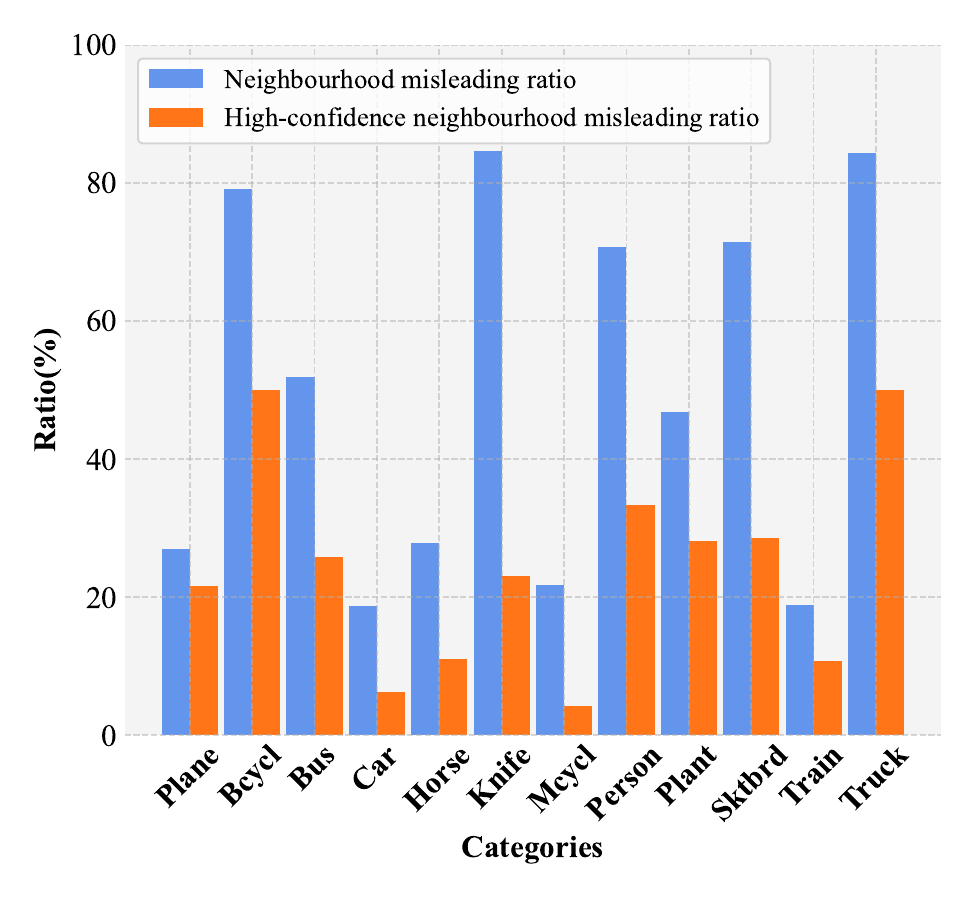}
    \label{fig:veri-a}
    
  \end{subfigure}
  \vspace{-0.7cm}
  \caption{{(Left) The pair-wise relation for sample $c$ only considers the affinity to sample $b$ in its neighborhood, but it fails to consider the high-order relation between sample $a$ and $c$, resulting in inaccurate predictions.} (Middle) Comparison of the pair-wise relation based method \cite{AaD} and our method on the accuracy of target samples' nearest neighbors having the correct predicted labels. {A higher accuracy indicates similar samples are well-clustered, which thereby demonstrates using high-order relations enables better clustering.} (Right) {``Neighborhood misleading ratio'' and ``High-confidence'' denote the mismatch between the predicted label and ground truth label of neighbors, and neighbors with high prediction confidence~\cite{lln2023source}.} Without involving the domain shift in optimization, the misleading ratio fluctuates among different categories, indicating the domain shift is not generally solved. These figures are validated on the VisDA dataset \cite{peng2017visda}. }
  \label{fig:veri}
  \vspace{-0.5cm}
\end{figure}

Deep learning methods for vision tasks, trained with a large number of training samples, can generalize well on the testing set with a similar data distribution~\cite{he2016deepresnet50,liang2020polytransform,liu2022adaptive,carion2020end}. However, their performance notably degrades when applied to an unseen data distribution due to the phenomenon of Domain Shift, \ie, differences in the data distribution between the source and target domains. Unsupervised Domain Adaptation (UDA) is a typical solution to this issue by transferring knowledge from the fully labeled source domain to the unlabeled target domain~\cite{DAcui2020towards,DAgong2012geodesic,DAlong2018transferable,SRDC,ADDA}. 
Nevertheless, traditional UDA methods require to access to the data of the source domain during training, which may be infeasible in real-world applications due to data privacy or intellectual property concerns~\cite{fang2022source,yu2023comprehensive}.

One emerging research direction, Source-free Unsupervised Domain Adaptation (SFDA), has recently been explored to address the above concerns and attracted increasing attention~\cite{G-SFDA,A2Net,shot,qu2022bmd,c-SFDA,zhang2023class,2023guiding}. The setting of SFDA is stricter and more challenging than UDA because the source data is unavailable, and only a pre-trained source model and target data are available. Under this setting, obtaining more domain knowledge depends on how to effectively exploit the underlying relation of these target samples. Although several methods attempt to solve this problem~\cite{3c-GAN,VDM-DA,NRC,DIPE,A2Net,zhang2023class}, most of them are developed based on the spirit of neighborhood clustering so that domain adaptation can be accomplished by exploring the neighborhood relation of target samples in feature space~\cite{qiu2021source,ma2021semi,li2022source,AaD,ICLR2024}. The intuition behind these methods is that similar target samples likely belong to the same semantic class and vice versa, and the sample relations in clusters can help model learn domain invariant knowledge. Despite promising results shown in these methods, they still have the following limitations: 
{\bf (1) Only pair-wise relations are considered in clustering.}
Since no prior knowledge about the source data is available, only considering the pair-wise sample relations may not adequately capture the underlying relations hidden in the target domain, {see Fig.~\ref{fig:veri}(Left).} This limitation results in failing to capture deeper structural information and makes the model easily distracted by outliers\footnote{Outlier denotes the sample that is wrongly predicted.}, which directly hinders the model from learning domain invariant knowledge, as seen in Fig.~\ref{fig:veri}(Middle).  
{\bf (2) Domain shift is not explicitly involved in clustering.}
Existing works focus on seeking the semantic relation of target samples and assume the domain shift can be reduced implicitly by only considering the semantic relation. This strategy cannot effectively address the domain shift problem, as it is not explicitly involved in the clustering process, thereby hindering clustering effectiveness, as shown in Fig.~\ref{fig:veri}(Right).

\begin{figure*}[t]
    \centering
    \includegraphics[width=0.95\linewidth]{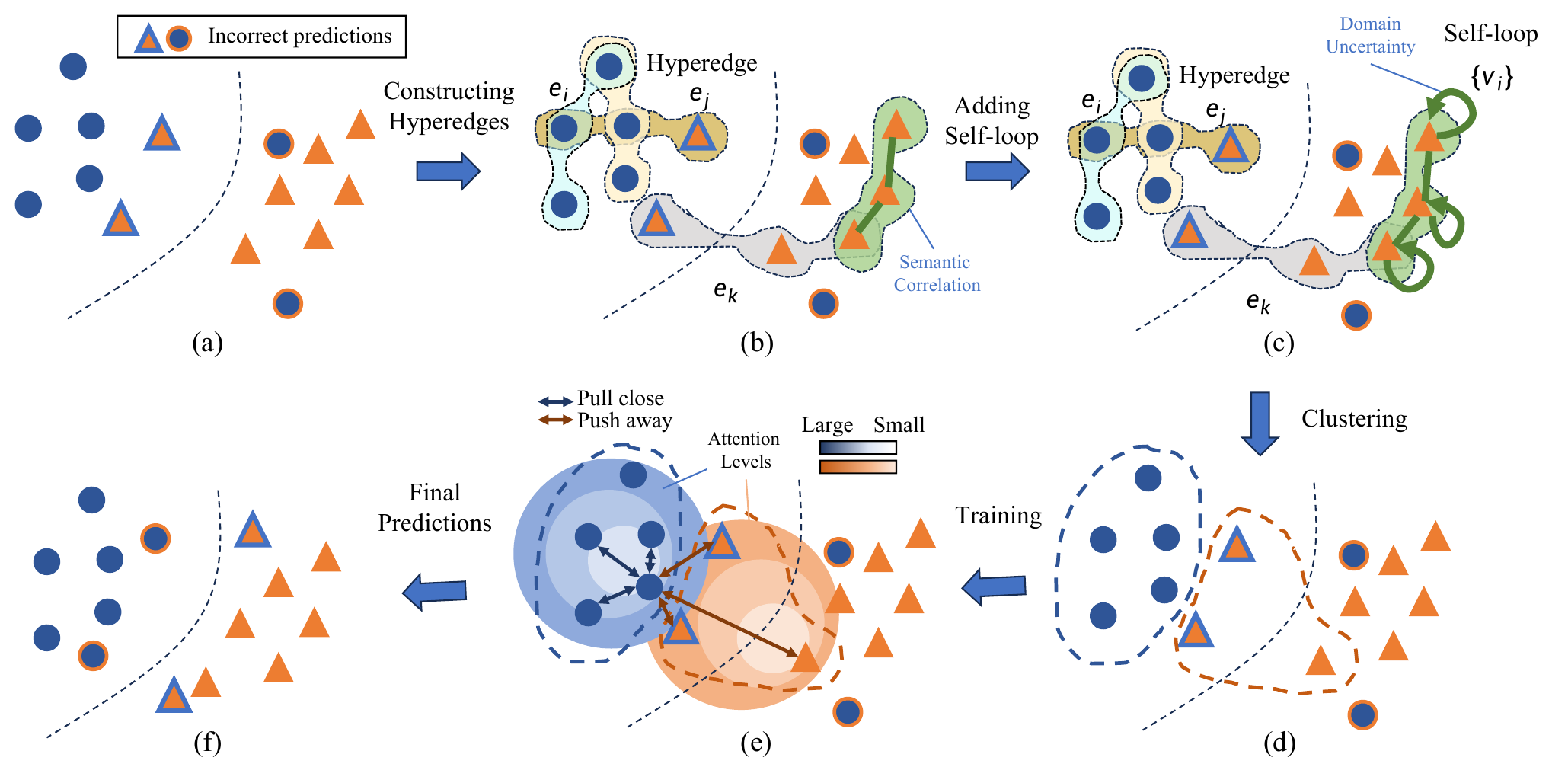}
    \vspace{-0.3cm}
  \caption{Overview of the proposed method {\em Hyper-SFDA}. (a) Initial results. (b) The hyperedges are constructed on the target domain to capture complicated higher-order neighborhood relations among multiple samples. (c) A self-loop strategy is proposed to consider the domain shift effect. (d) Clustering results by considering both hyperedges and self-loops. (e) After clustering, the model is trained using the proposed Adaptive Relation-based Objective, which pulls close samples in the same cluster and pushes away samples in different clusters with different attention levels. (f) Final results. See text for more details.}
  \label{fig:short2}
  \vspace{-0.6cm}
\end{figure*}

In this paper, we present a novel SFDA method called {\em Hyper-SFDA} to overcome the above limitations. Fig.~\ref{fig:short2} shows the overview of our method. Differing from existing approaches, our method explores the high-order neighborhood relations among multiple target samples instead of pair-wise relations while considering the domain shift phenomenon explicitly. Since high-order neighborhood relations can encapsulate the complex interplay among two or more target samples and little prior knowledge is used in the SFDA setting, this high-order neighborhood is the most valuable and handy resource that can aggregate more local grouping information and context. 
To capture the high-order relation, we propose a hypergraph learning method, which formulates the target samples as graph nodes and conducts hyperedges over the graph. To form a hyperedge, we treat each node as a centroid and seek its nearest neighbors based on their semantic similarity. 

To attach importance to the domain shift effect, we propose a novel self-loop strategy on the constructed hypergraph. This strategy involves creating self-loops on nodes to represent the domain uncertainty of corresponding samples. Domain uncertainty is closely tied to the domain shift problem, as it indicates the likelihood of samples belonging to the source domain or target domain. 
{By involving the self-loops in clustering, the samples with high domain uncertainty are drawn more attention, which leads to a comprehensive consideration of both semantic relations and domain shift re-calibration, ultimately improving the effectiveness of clusters.} We assess the domain uncertainty of samples based on their entropy values. The higher the uncertainty of a sample is, the larger the value of its self-loop is. 

Furthermore, we propose a new objective function on hyperedges using an adaptive learning scheme. The basic idea is to push close samples in a cluster and push away samples in different clusters by referring to the hyperedges. In particular, we assign ``soft'' attention levels for different samples, \ie, paying more attention to hard samples and vice versa. For example, the samples having large differences in the same cluster should be concerned more than others. This also holds for samples from different clusters. Therefore, we adaptively assign different weights to samples according to the semantic distance between the target sample and its nearest neighbors. In addition, to further mitigate the effect of noisy labels caused by domain shifts, we propose a regularization term that can instruct the prediction by accumulated knowledge from previous time stamps. 

Extensive experiments are conducted on several image datasets (Office-31 \cite{office31}, Office-home \cite{officehome}, VisDA \cite{peng2017visda}) and a 3D point cloud dataset (PointDA-10 \cite{pointdan}), to compare our method to the recent counterparts with the best results currently available. The results obtained show the superiority of our method on the SFDA problem. 

The contributions of this paper can be summarized as follows.
\begin{enumerate}
    \item [-] We formulate the source-free unsupervised domain adaptation (SFDA) as a hypergraph learning problem and explore the high-order neighborhood relations among target samples to excavate the underlying structural information.  
    \item [-] With the constructed hypergraph, we design a novel self-loop strategy to elegantly involve the domain shift into optimization. 
    \item[-] We describe an adaptive objective function to pull close and push away samples under inter-cluster and intra-cluster settings with different attention levels.
\end{enumerate}

\vspace{-0.6cm}
\section{Related Work}
\vspace{-0.2cm}
\noindent{\bf Unsupervised Domain Adaptation.}
Unsupervised Domain Adaptation (UDA) methods aim to transfer the knowledge from the fully labeled source domain to the unlabeled target domain. Generally, the UDA methods can be divided into several categories, ranging from using the minimization of distributional differences~\cite{77,147,151,198,224,225,CoVi,RAIN}, adversarial training~\cite{73,74,216} to clustering~\cite{CoDT,CAT,SRDC,249,COT}. The distributional differences are usually minimized using Maximum Mean Discrepancy (MMD) \cite{84} and Contrastive domain discrepancy (CCD) \cite{CCD}. In addition to minimizing distributional differences, domain adaptation can also be accomplished through domain adversarial approaches. DANN~\cite{74} and VRADA~\cite{189} effectively confuse domain classifiers by countering their gradients using a gradient inversion layer. More recently, Clustering-based methods have gained popularity, which can discover the correlation of samples between source and target domain and extract the domain invariant knowledge. For example, CoDT~\cite{CoDT} captures robust pseudo-labels to guide feature clustering by exploiting the complementary domain-shared features and target-specific features. CAT~\cite{CAT} achieves the goals of domain alignment and class-conditional alignment through a discriminative clustering loss and a clustering-based alignment loss.

\smallskip
\noindent{\bf Source-free Unsupervised Domain Adaptation.}
Source-free Unsupervised Domain Adaptation (SFDA) is a more challenging category of UDA, which requires accomplishing domain adaptation only with a pre-trained source model and unlabeled target data~\cite{21,22,23,U-SFAN,lln2023source,HCL,Feat-Mixup,c-SFDA,IJCV2023,zhang2023ICCV}. In the early stage, the methods ~\cite{26,28,36,37,VMP,DIPE,DMCD,Sub-Sup,3c-GAN,SFDA-DE,VDM-DA} focused on learning domain invariant representations to facilitate cross-domain adaptation. Specifically, the work of \cite{37} introduced an image generator to update target images to resemble source images and \cite{36} employed a GAN-based generator to simulate source data. In recent years, following the clustering spirit in general UDA, many clustering-based strategies have been proposed to solve the SFDA problem~\cite{ICLR2024,NRC,G-SFDA,93,CoWA-JMDS,shot++,AaD,98,zhang2023class,qu2022bmd,NRC++}. For example, NRC++~\cite{NRC++} introduced a local structural clustering strategy to encourage prediction consistency among nearest neighbors with high affinity. SF(DA)$^{2}$~\cite{ICLR2024} proposed a spectral neighborhood clustering (SNC) loss based on AaD~\cite{AaD} to identify partitions in the prediction space by spectral clustering. However, these methods only focused on pair-wise relations between samples and overlooked the domain shift issue in optimization, failing to extract the underlying structural information of target data. 

\vspace{-0.4cm}
\section{Methods}
\vspace{-0.2cm}
In contrast to the existing methods, our method explores the high-order neighborhood relation of target samples inspired by the essence of the hypergraph and introduces a self-loop strategy on the constructed hypergraph to involve the domain uncertainty of samples in the clustering process. Moreover, we describe a new objective function that formulates the relation of samples adaptively according to their similarity and whether they are in a cluster to further improve the discriminative ability.

\vspace{-0.2cm}
\subsection{Problem Setting of SFDA}
\vspace{-0.2cm}
Denote the source domain as ${\cal D}_{s}={\left\{\left(x_{i}^{s},y_{i}^{s} \right) \right\}_{i=1}^{N_{s}}}$, where $x_{i}^{s},y_{i}^{s}$ represents a source sample and its corresponding label, $N_{s}$ is the number of samples. Denote the target domain as ${\cal D}_{t}={\left\{x_{i}^{t} \right\}_{i=1}^{N_{t}}}$ containing $N_{t}$ unlabeled samples. Let the model network be ${\cal O}$, which consists of a feature extractor $f$ and a classifier $g$. Given an input sample $x$, the output of the feature extractor is denoted as $z=f(x)$, and the prediction vector (after softmax) of the classifier is denoted as $p = g(z)$. The objective of SFDA is to transfer the knowledge from the source domain to the target domain, by adapting a pre-trained source model ${\cal O}$ to the target domain ${\cal D}_{t}$, without accessing to the source domain data ${\cal D}_{s}$. Following previous works \cite{NRC,AaD,c-SFDA}, we explore this task mainly on the close-set setting, \ie, the source domain and target domain share the same label set. 

\vspace{-0.2cm}
\subsection{Exploring High-order Neighborhood Relation}
\label{subsec:3.2}
\vspace{-0.2cm}
\smallskip
\noindent{\bf Hypergraph Definition.}
The hypergraph is a graph structure with special edges that cover two more nodes. Let ${\cal G}=({\cal V},{\cal E},{\cal W})$ represent a hypergraph, where ${\cal V} = \{v_1, ..., v_n \}$ is the set of nodes, ${\cal E} = \{e_1,...,e_m\}$ is the set of hyperedges, and ${\cal W}$ is the affinity set corresponding to hyperedges. Specifically, each hyperedge $e \in {\cal E}$ consists of $k (k>2)$ nodes, and the degree of each hyperedge is $k = \sum_{v \in e} \mathbf{1}$. ${\cal W}(e)$ denotes the affinity associated with hyperedge $e$. 
Formally, the hypergraph ${\cal G}$ can be represented by a relation matrix ${\cal H}$ with a size $n \times m$, where each element ${\cal H}\left(v,e \right)=1$ if the node $v$ exists in the hyperedge $e$, otherwise ${\cal H}\left(v,e \right)=0$.
For each node $v$, we use ${\cal N}_{}(v)$ to denote its neighborhood which is a set containing nodes connected to $v$.

\smallskip
\noindent{\bf Hypergraph Meets SFDA.}
Given the target domain ${\cal D}_t$, we formulate a hypergraph structure based on the target samples. Specifically, we set each target sample $x^t$ as a node $v$, \ie, $v = x^t$ and form the exploration for samples relation given in the previous works as the proper building of hyperedges to effectively expose the underlying relation from a high-order perspective. We then seek the clusters for each target sample based on the hyperedges and use these clusters to drive the fine-tuning of model ${\cal O}$ on target domain ${\cal D}_t$. 


\smallskip
\noindent{\bf Hyperedge Generation.}
How to precisely select the nodes and measure the affinity of these nodes are fundamentally important in generating hyperedges. Specifically, we aim to generate hyperedges corresponding to every node, \ie, $n = m$. For each node $v_i$, we set it as an anchor node and find its $k-1$ nearest neighbors. These $k-1$ neighbors and the node $v_i$ together form a hyperedge $e_i$ with degree $k$, denoted as $e_i = \{v_i\} \cup \{v_{i_1}, ..., v_{i_{k-1}} \}$. To obtain the nearest neighbors, we measure the similarity between samples using their cosine similarity of features $z$ and then employ the KNN algorithm \cite{AaD} to select $k-1$ neighbors based on the calculated similarity. 

Given a hyperedge, we formulate its affinity by measuring the coherence of this hyperedge. Inspired by the work \cite{l1}, we describe the coherence using the relation between its anchor node and other nodes in the form of a linear combination. Assume that the feature of node $v_i$ can be reconstructed by a linear combination of its $k-1$ neighbors, and the coefficient of each neighbor in this combination indicates the relation of node $v_i$ to each neighbor. To obtain the affinity of a hyperedge $e_i$, we optimize the following objective function based on node $v_i$, as 
\begin{equation}
\begin{array}{c}
\mathop{\arg \min}\limits_{\mathbf{a}_i} \left\| {\cal F}({\cal N}_{k-1}(v_i), \mathbf{a}_i) - f\left(v_i \right)\right\|_2^2 + \alpha \left\| \mathbf{a}_i \right\|_2\\
\text{s.t.} \; \forall j, a_{i,j} \ge 0, a_{i,j} \in \mathbf{a}_i,
\end{array}
  \label{eq:important1}
\end{equation}
where $\mathbf{a}_i = \{a_{i,1}, ..., a_{i,k-1} \}$ is the coefficient vector for node $v_i$ and each element $a_{i,j}$ corresponds to the coefficient for $j$-th neighbor of node $v_i$. $\left\| \mathbf{a}_i \right\|_2$ is the regularization term to make the coefficient vector sparse and $\alpha$ is a balancing factor, while ${\cal N}_{k-1}(v_i) = \{v_{i_1}, ..., v_{i_{k-1}} \}$ is the set of $k-1$ neighbors of node $v_i$ obtained by KNN. In Eq.~\eqref{eq:important1}, ${\cal F}({\cal N}_{k-1}(v_i), \mathbf{a}_i)$ denotes the linear combination operation as
\begin{equation}
\begin{array}{c}
{\cal F}({\cal N}_{k-1}(v_i), \mathbf{a}_i) = \sum_{j}^{k-1} a_{i,j} \cdot f(v_{i_j}).
\end{array}
  \label{eq:important2}
\end{equation}
For the hyperedge $e_i = \{v_i\} \cup \{v_{i_1}, ..., v_{i_{k-1}} \}$, its affinity can be represented by the vector $\mathbf{a}_i$ as 
\begin{equation}
\begin{array}{rl}
{\cal W}(e_i) = & \{1, a_{i,1}, ..., a_{i,k-1} \} \\
= & \{1 \} \cup \mathbf{a}_i,
\end{array}
  \label{eq:important3}
\end{equation}
where $1$ is the fixed coefficient for the anchor node $v_i$. We perform this optimization over all nodes, and each node $v_i$ corresponds to a hyperedge $e_i$ with its own affinity as ${\cal W}(e_i)$. Note in our solution, the number of nodes is equal to the number of hyperedges.


\vspace{-0.2cm}
\subsection{Handling Domain Shift by Self-loop}
\label{subsec:3.3}
\vspace{-0.2cm}

Generating the hyperedge using the above strategies does not consider the domain shift effect in clustering. This means that all samples are assumed to have equal domain uncertainty. However, different samples should have different levels of domain uncertainty. For example, the samples distributed around the boundary between the source and target domain should have high uncertainty, which requires more attention during optimization. Therefore, the generation of hyperedges should consider this uncertainty along with semantic relations in order to alleviate the domain shift problem. 
As such, we develop a self-loop for each target sample to indicate its domain uncertainty. Specifically, we estimate this domain uncertainty using the entropy value based on the pseudo-labels from the classifier $g$. Samples with high entropy indicate high uncertainty and are viewed as challenging samples with large domain shifts, whereas samples with low entropy are relatively simple, enabling the model to concentrate on samples with substantial domain shifts. 

Denote ${\cal E}_s = \{ \{v_1\}, ..., \{v_n\} \}$ as the self-loops for nodes. By adding self-loops, the hypergraph can be updated as ${\cal G}=({\cal V},{\cal E} \cup {\cal E}_s,{\cal W} \cup {\cal W}_s)$, where ${\cal W}_s$ is the affinity set of self-loops. To obtain the affinity of a self-loop, the most straightforward way is to calculate its entropy using the prediction vector $p$. However, solely relying on one node may suffer from the inner deviation of this node. Therefore, we consider its neighbors for representation to mitigate the errors. Specifically, given a node $v_i$, we find its corresponding hyperedge $e_i$ and average the prediction vectors of the other $k-1$ nodes in this hyperedge. Then we calculate the entropy based on the averaged prediction vector and employ the exponential function for normalization. The rationale for using the exponential function is that it can assign a larger weight to samples with high entropy, allowing us to prioritize the samples likely to have shifted. The affinity of the self-loop is calculated as 
\begin{equation}
\begin{array}{c}
{\cal W}_s(\{v_i\}) = \exp (\phi(\bar{p}_i)), 
\end{array}
  \label{eq:important4}
\end{equation}
where $\bar{p}_i$ is the averaged prediction vector, defined as
\begin{equation}
\begin{array}{c}
\bar{p}_i = \frac{1}{k-1}\sum_{v_j \in e_i / v_i} g(f(v_j)),
\end{array}
  \label{eq:important5}
\end{equation}
where $e_i / v_i$ denotes the nodes in $e_i$ except $v_i$, $\phi(\cdot)$ denotes the calculation of entropy as
\begin{equation}
\begin{array}{c}
\phi(\bar{p}_i) = \frac{1}{\log |C|}\sum_{c \in C} -\bar{p}_i^c \log \bar{p}_i^c,
\end{array}
  \label{eq:important6}
\end{equation}
where $C$ is the set of class categories. 

Based on each hyperedge obtained above, we add self-loops into every node to form a new hyperedge. For the nodes in hyperedge $e_i$, the affinity of their self-loops can be defined as 
\begin{equation}
\begin{array}{c}
\mathbf{b}_i = \{ {\cal W}_s(\{v_i\}) \} \cup \{{\cal W}_s\{v_{i_1}\}, ..., {\cal W}_s\{v_{i_{k-1}}\} \}
\end{array}
  \label{eq:important7}
\end{equation}
After adding self-loops, the affinity of hyperedge $e_i$ can be defined as ${\cal W}(e_i) = {\cal W}(e_i) + \mathbf{b}_i$. Thus the affinity of all hyperedges is a set of $\{ {\cal W}(e_1), ..., {\cal W}(e_m) \}$.

\vspace{-0.2cm}
\subsection{Clustering and Training}
\label{subsec:3.4}
\vspace{-0.2cm}
Based on the generated hyperedges, we search for the high-order nearest neighbors for each node. Then we design objective functions to guide the tuning of the model based on the explored relation from these clusters. The training procedure is illustrated in Alg.~\ref{alg}.

\smallskip
\noindent{\bf Clustering based on High-order Relations.}
To find the cluster for each node, we update the relation matrix ${\cal H}$ with representations calculated in Eq.~\eqref{eq:important1} instead of the fixed value $1$ or $0$. Mathematically, the element value of ${\cal H}$ can be updated as
\begin{equation}
\begin{aligned}
{\cal H}(v_i,e_j) = \begin{cases}
{\cal W}\left(e_j \right) |_{v_i}, &  v_i \in e_j \\
0,& v_i \not \in e_j \\
\end{cases}
\end{aligned}
  \label{eq:important8}
\end{equation}
where ${\cal W}\left(e_j \right) |_{v_i}$ means to pick the element in ${\cal W}\left(e_i \right)$ corresponding to node $v_i$. Given the relation matrix ${\cal H}$, we can obtain the relation of all nodes to all hyperedges with each row represents the relation of this node to all hyperedges, which reflects the knowledge of this node correlating with the hypergraph, \ie, the target domain. Thus it can be viewed as a compact representation ($1 \times m$) for this node. Then we perform KNN based on the representation of this node and find the top-$h$ neighbors to form a cluster. 

To reduce the clustering cost, we compact the representation for each node by projecting the ${\cal H}$ from $n \times m$ to $n \times m' (m' < m)$ (\eg, PCA). Therefore, for each node $v_i$, we can obtain a set of neighbors as ${\cal A}_i$ and regard the rest nodes as a background set ${\cal B}_i$.

\smallskip
\noindent{\bf Adaptive Relation-based Objectives.}
Based on the above cluster results, we design adaptive relation-based objectives to pull close the samples in a cluster and push away samples in different clusters, while assigning ``soft'' attention levels to different samples. Specifically, we examine their prediction similarity and adaptively assign different weights to these samples according to the following criteria: 1) In a cluster, samples should be coherent. Thus we pay more attention to the hard samples that have notable discrepancies, facilitating the aggregation of samples into the same category. 2) For different clusters, samples should have clear differences. Thus we emphasize the refinement of hard samples having small distances and push those samples away to boost the discriminative ability of the model. Our adaptive relation-based objective loss can be defined as
\begin{equation}
\begin{aligned}
{\cal L}_{ada}=-\sum_{j\in {\cal A}_{i}}\left(1-d_{ij}^{\gamma}\right)p_{i}^{\top}p_{j} + \lambda\sum_{k\in {\cal B}_{i}}\left(1-d_{ik}^{\gamma}\right)p_{i}^{\top}p_{k},
\end{aligned}
  \label{eq:important9}
\end{equation}
where $d_{ij}$ denotes the normalized Euclidean distance between the $p_i$ and $p_j$, and $\gamma$ is a scale factor controlling the magnitude of distance. The first term adaptively pulls close samples in the same cluster and the second term adaptively pushes away samples in different clusters. $\lambda$ is the weighting factor to balance these two loss terms, which is set to ${\lambda=(1+10 \cdot \frac{\text{iter}}{\text{maxiter}})^{-\beta}}$ dynamically as in \cite{AaD}.

Moreover, to further improve the performance, we describe a regularization term that aims to 
prevent the model from overly focusing on incorrect predictions. Inspired by the early training phenomenon in \cite{lln2023source}, we use a weighted moving average method~\cite{Weight-averaged}, which accumulates knowledge of the predictions at previous training time stamps, in order to provide more precise instruction for current prediction. For $t$-th time stamp (iteration), the target prediction is defined as $q^{(t)}_{i}= \delta  q^{(t-1)}_{i} + \left(1-\delta \right)p^{(t)}_{i}.$
The initial state of $q^{(0)}_i$ is set to 0 and $\delta$ is the weight factor. To make the current prediction approach the target prediction, we use KL divergence to measure their difference, as ${\cal L}_{reg} = \text{KL}(q_i || p_i)$.
The overall objective function is the combination of these two loss terms, where $\eta$ is a trade-off hyperparameter.
\begin{equation}
\begin{aligned}
{\cal L} = {\cal L}_{ada} + \eta{\cal L}_{reg}.
\end{aligned}
  \label{eq:important10}
\end{equation}

\begin{algorithm}[t]
    \footnotesize
    \caption{\small {Overall training procedure of Hyper-SFDA}}
    \label{alg}
    \begin{algorithmic}
    \Require Target domain ${\cal D}_t$, Source model ${\cal O}$, Total training iterations $T$, Interval of updating hypergraph $T_{in}$
    \Ensure Fine-tuned model ${\cal O}$
    \For{$t=0 \rightarrow  T$}
    \If {$t \; \% \; T_{in} = 0$}
    \State Constructing hypergraph ${\cal G}$
    \State Generating hyperedges ${\cal E}$ and calculating affinity set ${\cal W} \cup {\cal W}_s$
    \EndIf
    \State Training batch ${\cal V}_b$ $\sim$ Target domain ${\cal D}_t$
    \For {node $v_i$ $\sim$ Training batch ${\cal V}_b$}
    \State Performing clustering based on node $v_i$
    \EndFor
    \State Training model ${\cal O}$ on ${\cal V}_b$ using objective in Eq.~\eqref{eq:important10}
    \EndFor
\end{algorithmic}
\end{algorithm}


\vspace{-0.8cm}
\section{Experiments}
\label{sec:experiment}
\vspace{-0.2cm}
\subsection{Experimental Setup}
\vspace{-0.2cm}
\textbf{Datasets.} We evaluate our method on three commonly used 2D image benchmark datasets, \textbf{Office-31}~\cite{office31}, \textbf{Office-Home}~\cite{officehome} and \textbf{VisDA}~\cite{peng2017visda}, and one challenging 3D point cloud recognition dataset \textbf{PointDA-10}~\cite{pointdan}. The Office-31 dataset contains $3$ domains of  Amazon, Webcam, and DSLR with $31$ classes and $4652$ images. The Office-Home dataset contains $4$ domains of Real, Clipart, Art, and Product with $65$ classes and a total of $15,500$ images. VisDA is a large-scale dataset with $12$ classes for both synthetic and real object recognition tasks, containing $152,000$ synthetic images in the source domain and $55,000$ real object images in the target domain. PointDA-10 is a 3D point cloud benchmark dataset designed for a domain adaptation, with $3$ domains and $10$ classes, denoted as ModelNet-10, ShapeNet-10, and ScanNet-10. It contains a total of $27,700$ training images and $5,100$ test images.

\smallskip
\noindent \textbf{Implementation details.} Following the previous works \cite{zhang2023class,c-SFDA,qu2022bmd}, we used ResNet-50~\cite{he2016deepresnet50} as the backbone network on Office-31 and Office-Home dataset, and use ResNet-101 on VisDA dataset for a fair comparison. For the PointDA-10 dataset, we use PointNet as in ~\cite{qi2017pointnet}. 
In the training stage, we employ SGD optimizer with a momentum of $0.9$ for the 2D image datasets and use Adam optimizer for the PointDA-10 dataset. The batch size for all datasets is set to $64$. The starting learning rate for 2D image datasets is set to $1 \times 10^{-3}$, and the one for the PointDA-10 dataset is set to $1 \times 10^{-6}$. We train $50$ epochs for Office-31 and train $40$ epochs for Office-Home while $35$ epochs for VisDA, and $100$ epochs for PointDA-10. To construct hyperedges, we set the degree $k = 6$ and update the hypergraph structure every $T_{in}=50$ iterations. {For Eq.~\eqref{eq:important1}, we set $\alpha=2$.} To cluster the samples, we consider $h=3$ nearest neighbors. For ${\cal L}_{reg}$, we set $\delta$, $\eta$ to $0.8$, $2$, respectively. More training and hyperparameter details can be found in the supplementary material.  

\vspace{-0.2cm}
\subsection{Results}
\vspace{-0.2cm}
The results of our method and other existing benchmark counterparts on Office-Home, Office-31, VisDA, and PointDA-10 datasets are shown in Table~\ref{table_ip:1} to ~\ref{table_ip:3}. In each table, we show the results of each task and their average accuracy over all tasks (Avg). The best results are marked in bold. SF denotes Source-Free unsupervised domain adaptation and $\times$ means the method requires access to source domain data during domain adaptation, while $\checkmark$ means the method falls into the SFDA setting. 

\begin{table*}[t]
 \centering
 \footnotesize
 \caption{Accuracy (\%) of the methods on Office-Home dataset.}
 \vspace{-0.3cm}
 \label{table_ip:1}
 \setlength{\tabcolsep}{0.85mm}{
\scalebox{0.7}{
  \begin{tabular}{lccccccccccccccc}
   \hline
 Method & SF & A→C & A→P & A→R & C→A & C→P & C→R & P→A & P→C & P→R & R→A & R→C & R→P & Avg \\       
   \hline                                                                          CoVi (ECCV’22)~\cite{CoVi} & × & 58.5 & 78.1 & 80.0 & 68.1 & 80.0 & 77.0 & 66.4 & 60.2 & 82.1 & \textbf{76.6} & 63.6 & 86.5 & 73.1 \\
RAIN (IJCAI’23)~\cite{RAIN} & × & 57.0 & 79.7 & 82.8 & 67.9 & 79.5 & 81.2 & 67.7 & 53.2 & 84.6 & 73.3 & 59.6 & 85.6 & 73.0 \\
COT (CVPR’23)~\cite{COT} & × & 57.6 & 75.2 & 83.2 & 67.8 & 76.2 & 75.7 & 65.4 & 56.2 & 82.4 & 75.1 & 60.7 & 84.7 & 71.7 \\
   \hline                                                                                        SHOT (ICML’20)~\cite{shot} & \checkmark & 57.1 & 78.1 & 81.5 & 68.0 & 78.2 & 78.1 & 67.4 & 54.9 & 82.2 & 73.3 & 58.8 & 84.3 & 71.8 \\         
NRC (NeurIPS’21)~\cite{NRC} & \checkmark & 57.7 & 80.3 & 82.0 & 68.1 & 79.8 & 78.6 & 65.3 & 56.4 & 83.0 & 71.0 & 58.6 & 85.6 & 72.2 \\
DIPE (CVPR’22)~\cite{DIPE} & \checkmark & 56.5 & 79.2 & 80.7 & 70.1 & 79.8 & 78.8 & 67.9 & 55.1 & 83.5 & 74.1 & 59.3 & 84.8 & 72.5 \\
CoWA-JMDS (ICML’22)~\cite{CoWA-JMDS} & \checkmark & 56.9 & 78.4 & 81.0 & 69.1 & 80.0 & 79.9 & 67.7 & 57.2 & 82.4 & 72.8 & 60.5 & 84.5 & 72.5 \\
VMP (NeurIPS’22)~\cite{VMP} & \checkmark & 57.9 & 77.6 & 82.5 & 68.6 & 79.4 & 80.6 & 68.4 & 55.6 & 83.1 & 75.2 & 59.6 & 84.7 & 72.8 \\
D-MCD (AAAI’22)~\cite{DMCD} & \checkmark & 59.4 & 78.9 & 80.2 & 67.2 & 79.3 & 78.6 & 65.3 & 55.6 & 82.2 & 73.3 & 62.8 & 83.9 & 72.2 \\
AaD (NeurIPS’22)~\cite{AaD} & \checkmark & 59.3 & 79.3 & 82.1 & 68.9 & 79.8 & 79.5 & 67.2 & 57.4 & 83.1 & 72.1 & 58.5 & 85.4 & 72.7 \\
Sub-Sup (ECCV’22)~\cite{Sub-Sup} & \checkmark & 61.0 & 80.4 & 82.5 & 69.1 & 79.9 & 79.5 & 69.1 & 57.8 & 82.7 & 74.5 & \textbf{65.1} & 86.4 & 74.0 \\
BMD (ECCV’22)~\cite{qu2022bmd} & \checkmark & 58.1 & 79.7 & 82.6 & 69.3 & 81.0 & 80.7 & 70.8 & 57.6 & 83.6 & 74.0 & 60.0 & 85.9 & 73.6 \\
U-SFAN (ECCV’22)~\cite{U-SFAN} & \checkmark & 57.8 & 77.8 & 81.6 & 67.9 & 77.3 & 79.2 & 67.2 & 54.7 & 81.2 & 73.3 & 60.3 & 83.9 & 71.9 \\
TPDS (IJCV’23)~\cite{IJCV2023} & \checkmark & 59.3 & 80.3 & 82.1 & 70.6 & 79.4 & 80.9 & 69.8 & 56.8 & 82.1 & 74.5 & 61.2 & 85.3 & 73.5 \\
NRC++ (TPAMI’23)~\cite{NRC++} & \checkmark & 57.8 & 80.4 & 81.6 & 69.0 & 80.3 & 79.5 & 65.6 & 57.0 & 83.2 & 72.3 & 59.6 & 85.7 & 72.5 \\
CREL (CVPR’23)~\cite{zhang2023class} & \checkmark & \textbf{62.8} & 82.0 & \textbf{84.3} & 70.9 & 80.8 & \textbf{82.6} & 70.0 & \textbf{61.1} & 83.6 & 76.2 & \textbf{65.1} & 87.0 & 75.5 \\
C-SFDA (CVPR’23)~\cite{c-SFDA} & \checkmark & 60.3 & 80.2 & 82.9 & 69.3 & 80.1 & 78.8 & 67.3 & 58.1 & 83.4 & 73.6 & 61.3 & 86.3 & 73.5 \\
LLN (ICLR’23)~\cite{lln2023source} & \checkmark & 58.4 & 78.7 & 81.5 & 69.2 & 79.5 & 79.3 & 66.3 & 58.0 & 82.6 & 73.4 & 59.8 & 85.1 & 72.6 \\
Co-learn (ICCV’23)~\cite{zhang2023ICCV} & \checkmark & 57.7 & 80.4 & 83.3 & 70.1 & 80.1 & 80.6 & 66.6 & 55.5 & 84.1 & 72.1 & 57.6 & 84.3 & 72.7 \\
\rowcolor{mygray}{\bf Ours} & \checkmark & 62.6 & \textbf{82.4} & 84.2 & \textbf{73.1} & \textbf{82.6} & 82.4 & \textbf{72.0} & 60.6 & \textbf{85.0} & 76.4 & 63.0 & \textbf{87.6} & \textbf{76.0} \\ 
   \hline
 \end{tabular}}}
 \vspace{-0.3cm}
\end{table*}


\begin{table*}[t]
 \centering
 \footnotesize
 \caption{Accuracy (\%) of the methods on Office-31 (left columns) and VisDA dataset (rightmost column).}
 \vspace{-0.3cm}
 \label{table_ip:2}
 \setlength{\tabcolsep}{1.2mm}{
 \scalebox{0.75}{
  \begin{tabular}{lcccccccccc}
   \hline
\multirow{2}{*}{Method} & \multirow{2}{*}{SF} & \multicolumn{6}{c}{Office-31} &  &  & VisDA \\ \cline{3-9} \cline{11-11} 
 &  & A→D & A→W & D→W & W→D & D→A & W→A & Avg &  & S → R \\ \cline{1-9} \cline{11-11} 
   \hline                                  
CoVi (ECCV’22)~\cite{CoVi} & × & 98.0 & 97.6 & \textbf{99.3} & 100.0 & 77.5 & 78.4 & 91.8 &  & 88.5   \\
RAIN (IJCAI’23)~\cite{RAIN} & × & 93.8 & 88.8 & 96.8 & 99.5 & 75.5 & 76.7 & 88.5 &  & 82.7   \\
COT (CVPR’23)~\cite{COT} & × & 96.1 & 96.5 & 99.1 & 100.0 & 76.7 & 77.4 & 91.0 &  & 87.1  \\  \hline
                
SHOT (ICML’20)~\cite{shot} & \checkmark & 94.0 & 90.1 & 98.4 & 99.9 & 74.7 & 74.3 & 88.6 &  & 82.9   \\
HCL (NeurIPS’21)~\cite{HCL} & \checkmark & 90.8 & 91.3 & 98.2 & 100.0 & 72.7 & 72.7 & 87.6 &  & 83.5   \\
A$^{2}$Net (ICCV’21)~\cite{A2Net} & \checkmark & 94.5 & 94.0 & 99.2 & 100.0 & 76.7 & 76.1 & 90.1 &  & 84.3   \\
SHOT++ (TPAMI’21)~\cite{shot++} & \checkmark & 94.3 & 90.4 & 98.7 & 99.9 & 76.2 & 75.8 & 89.2 &  & 87.3   \\
NRC (NeurIPS’21)~\cite{NRC} & \checkmark & 96.0 & 90.8 & 99.0 & 100.0 & 75.3 & 75.0 & 89.4 &  & 85.9   \\
D-MCD (AAAI’22)~\cite{DMCD} & \checkmark & 94.1 & 93.5 & 98.8 & 100.0 & 76.4 & 76.4 & 89.9 &  & 87.5  \\
DIPE (CVPR’22)~\cite{DIPE} & \checkmark & 96.6 & 93.1 & 98.4 & 99.6 & 75.5 & 77.2 & 90.1 &  & 83.1   \\
CoWA-JMDS (ICML’22)~\cite{CoWA-JMDS} & \checkmark & 94.4 & 95.2 & 98.5 & 100.0 & 76.2 & 77.6 & 90.3 &  & 86.9   \\
Feat-Mixup (ICML’22)~\cite{Feat-Mixup} & \checkmark & 94.6 & 93.2 & 98.9 & 100.0 & 78.3 & \textbf{78.9} & 90.7 &  & 87.8   \\
SFDA-DE (CVPR’22)~\cite{SFDA-DE} & \checkmark & 96.0 & 94.2 & 98.5 & 99.8 & 76.6 & 75.5 & 90.1 &  & 86.5   \\
BMD (ECCV’22)~\cite{qu2022bmd} & \checkmark & 96.2 & 94.2 & 98.0 & 100.0 & 76.0 & 76.0 & 90.1 &  & 88.7   \\
Sub-Sup (ECCV’22)~\cite{Sub-Sup} & \checkmark & 95.6 & 94.6 & 99.2 & 99.8 & 77.0 & 77.7 & 90.7 &  & 88.2   \\
U-SFAN (ECCV’22)~\cite{U-SFAN} & \checkmark & 94.2 & 92.8 & 98.0 & 99.0 & 74.6 & 74.4 & 88.8 &  & 82.7   \\
AaD (NeurIPS’22)~\cite{AaD} & \checkmark & 96.4 & 92.1 & 99.1 & 100.0 & 75.0 & 76.5 & 89.9 &  & 88.0   \\
LLN (ICLR’23)~\cite{lln2023source} & \checkmark & - & - & - & - & - & - & - &  & 86.4   \\
TPDS (IJCV’23)~\cite{IJCV2023} & \checkmark & 97.1 & 94.5 & 98.7 & 99.8 & 75.7 & 75.5 & 90.2 &  & 87.6  \\
CREL (CVPR’23)~\cite{zhang2023class} & \checkmark & 95.8 & 95.1 & 99.0 & 100.0 & 76.6 & 78.3 & 90.8 &  & 89.1   \\
NRC++ (TPAMI’23)~\cite{NRC++} & \checkmark & 95.9 & 91.2 & 99.1 & 100.0 & 75.5 & 75.0 & 89.5 &  & 88.1   \\
Co-learn (ICCV’23)~\cite{zhang2023ICCV} & \checkmark & 96.6 & 92.5 & 98.9 & 99.8 & 77.3 & 76.6 & 90.3 &  & 88.2   \\
C-SFDA (CVPR’23)~\cite{c-SFDA} & \checkmark & 96.2 & 93.9 & 98.8 & 99.7 & 77.3 & 77.9 & 90.5 &  & 87.8  \\
SF(DA)$^{2}$ (ICLR’24)~\cite{ICLR2024} & \checkmark & 95.8 & 92.1 & 99.0 & 99.8 & 75.7 & 76.8 & 89.9 &  & 88.1   \\
\rowcolor{mygray}{\bf Ours}  & \checkmark & \textbf{98.4} & \textbf{98.2} & 99.1 & \textbf{100.0} & \textbf{78.6} & 78.7 & \textbf{92.2} &  & \textbf{89.6}  \\
 \hline  
\end{tabular}}}
\vspace{-0.3cm}
\end{table*}


\begin{table*}[t]
 \centering
 \footnotesize
 \caption{Accuracy (\%) of the methods on PointDA-10 dataset.}
 \vspace{-0.3cm}
 \label{table_ip:3}
 \setlength{\tabcolsep}{1.2mm}{
 \scalebox{0.75}{
  \begin{tabular}{lcccccccc}
   \hline
 Method & SF & M→SC & M→SH & SC→M & SC→SH & SH→M & SH→SC & Avg \\    
   \hline                                                                          ADDA (CVPR’17)~\cite{ADDA} & × & 30.5 & 61.0 & 48.9 & 51.1 & 40.4 & 29.3 & 43.5 \\
MCD (CVPR’18)~\cite{MCD} & × & 31.0 & 62.0 & 46.8 & 59.3 & 41.4 & 31.3 & 45.3 \\
PointDAN (NeurIPS’19)~\cite{pointdan} & × & 33.0 & 64.2 & 49.1 & 64.1 & 47.6 & 33.9 & 48.7 \\ 
   \hline                                                        
SHOT (ICML’20)~\cite{shot} & \checkmark & 31.8 & 62.1 & 67.6 & 56.9 & 75.8 & 24.3 & 53.1 \\
VDM (Arxiv’21)~\cite{VDM-DA} & \checkmark & 30.9 & 58.4 & 45.3 & 61.8 & 61.0 & \textbf{40.8} & 49.7 \\
NRC (NeurIPS’21)~\cite{NRC} & \checkmark & 25.8 & 64.8 & 70.1 & 68.1 & 59.8 & 26.9 & 52.6 \\
AaD (NeurIPS’22)~\cite{AaD} & \checkmark & 34.6 & 69.6 & 68.0 & 66.6 & 67.7 & 28.8 & 55.9 \\
BMD (ECCV’22)~\cite{qu2022bmd} & \checkmark & 32.8 & 66.1 & 75.0 & 62.0 & \textbf{81.5} & 24.4 & 57.0 \\
CREL (CVPR’23)~\cite{zhang2023class} & \checkmark & 26.0 & 42.2 & 59.9 & 55.0 & 51.8 & 25.6 & 43.4 \\
NRC++ (TPAMI’23)~\cite{NRC++} & \checkmark & 27.6 & 67.2 & 74.5 & 71.2 & 60.2 & 30.4 & 55.1 \\
SF(DA)$^{2}$ (ICLR’24)~\cite{ICLR2024} & \checkmark & \textbf{35.5} & 70.3 & 70.4 & 69.2 & 68.3 & 29.0 & 57.1 \\
\rowcolor{mygray}{\bf Ours}  & \checkmark & 33.2 & \textbf{71.9} & \textbf{79.4} & \textbf{73.9} & 77.9 & 27.5 & \textbf{60.6} \\ 
   \hline
 \end{tabular}}}
 \vspace{-0.5cm}
\end{table*}

\smallskip 
\noindent \textbf{Office-Home.} Following previous works \cite{NRC++,NRC,AaD,lln2023source}, our method is compared to several state-of-the-art DA methods and SFDA methods ranging from 2020 to 2023. 
As shown in Table~\ref{table_ip:1}, our method achieves the best results on average, $76.0\%$, which outperforms all DA methods and surpasses the most recent SFDA methods C-SFDA \cite{c-SFDA}, LLN \cite{lln2023source} and Co-learn \cite{zhang2023ICCV} by $2.5\%,3.4\%,3.3\%$. It can also be seen that our method can outperform others in six out of twelve tracks, which demonstrates the general effectiveness of our method on this dataset.  

\smallskip 
\noindent \textbf{Office-31 and VisDA.} 
Table~\ref{table_ip:2} shows the performance of several methods on Office-31 and VisDA datasets. We evaluate more counterparts that are specifically designed for these two datasets. The results reveal that our method outperforms other methods even if they are not under the SFDA setting. On the Office-31 dataset, our method can achieve the best performance on four out of six tracks, demonstrating the effectiveness of our method.  

\smallskip 
\noindent \textbf{PointDA-10.} Table~\ref{table_ip:3} shows the performance of our methods compared with others on the PointDA-10 dataset. Since this dataset is for 3D point cloud recognition, only a few methods have reported their performance on it. It can be seen that our method largely improves the performance and achieves the best performance, $60.6\%$, surpassing the second-best method SF(DA)$^{2}$~\cite{ICLR2024} by $3.5\%$. {CREL~\cite{zhang2023class} is designed specifically for images, which is difficult to be adapted to PointDA-10. For comparison, we reproduce its losses on the baseline model, showing that it does not perform well in this task.} This experiment corroborates that our method is not only effective in general 2D images but also in the 3D point cloud recognition tasks.



\vspace{-0.2cm}
\subsection{Ablation Study}
\vspace{-0.2cm}

\smallskip 
\noindent{\textbf{Effect of Each Component.}} In this experiment, we investigate the effect of using the newly proposed components, which are the high-order neighborhood relation, self-loop strategy, and adaptive relation-based loss respectively. This experiment is validated on the Office-31 dataset. As shown in Table~\ref{table_ip:4}, the first row is the baseline performance without using any of these components, which is degraded to the AaD method~\cite{AaD}. By exploring the high-order information, the performance is improved by $0.7\%$. By adding the self-loop into hyperedges, we improve the performance by $1.1\%$. After adding the adaptive relation-based loss, the performance is further improved by $0.5\%$. 


\begin{table}[t]
    \centering
    \begin{minipage}[t]{0.33\textwidth}
        \footnotesize
        \caption{Effect of each component of our method. }
        \label{table_ip:4}
         \setlength{\tabcolsep}{1.2mm}{
         \scalebox{0.7}{
          \begin{tabular}{ccc|c}
           \hline
        High & Self & Adaptive & \multirow{2}{*}{Avg} \\  
        -order & -loop & Relation & \\
           \hline                                 
           × & × & × & 89.9 \\
        \checkmark & × & × & 90.6 \blue{(+0.7)}\\
        \checkmark & \checkmark & × & 91.7 \blue{(+1.1)}\\
        \checkmark & \checkmark & \checkmark & {\bf 92.2} \blue{(+0.5)} \\
           \hline                               
         \end{tabular}}}
        
    \end{minipage}
    \hfill
    \begin{minipage}[t]{0.3\textwidth}
        \centering
         \footnotesize
         \caption{Runtime analysis on Office-31.}
         \label{table_ip:5}
         \setlength{\tabcolsep}{1.2mm}{
         \scalebox{0.8}{
        	\begin{tabular}{c|c|c}
        		\hline
        		Methods & Runtime(s) & Acc (\%) \\ \hline
                    NRC~\cite{NRC} &   229 & 87.9 \\ \hline
        		AaD~\cite{AaD} &   414 & 92.6 \\\hline
        		LLN~\cite{lln2023source} &   330 & 92.2 \\ \hline
        		Ours            &   405 & 98.1 \\ \hline
        	\end{tabular}}}
        
    \end{minipage}
    \hfill
    \begin{minipage}[t]{0.3\textwidth}
        \centering
         \footnotesize
         \caption{Batch sizes analysis on Office-31.}
         \label{table_ip:6}
         \setlength{\tabcolsep}{1.2mm}{
         \scalebox{0.8}{
            \begin{tabular}{c|c|c|c}
        \hline
        \multirow{2}{*}{Methods} & \multicolumn{3}{c}{Batch Size} \\ 
        \cline{2-4}
                & 32   & 64   & 128 \\
        \hline
        NRC~\cite{NRC} & 90.2 & 90.8 & 90.1 \\ \hline
        AaD~\cite{AaD} & 93.3 & 92.6 & 91.5 \\ \hline
        LLN~\cite{lln2023source} & 91.8 & 92.2 & 91.8 \\ \hline
        Ours                  & 94.3 & 98.1 & 98.6 \\
        \hline
        \end{tabular}}}
        
    \end{minipage}
\end{table}



\smallskip
\noindent \textbf{Runtime Analysis.}
{Table~\ref{table_ip:5} shows the training time and accuracy of different methods on Office-31 (A$\rightarrow$W) under the same setting. Compared to the recent LLN~\cite{lln2023source} and AaD~\cite{AaD}, our method has comparable computation complexity but achieves better performance. 
Note that LLN, AaD, and our method utilize both intra-cluster and inter-cluster relations, whereas NRC~\cite{NRC} only considers the intra-cluster relations, leading to faster training but degraded performance. }

\smallskip
\noindent \textbf{Sensitivity of Batch Size.} {Table~\ref{table_ip:6} shows the performance of our method compared to the pair-wise methods using different batch sizes on Office-31 (A$\rightarrow$W). As observed, our method improves the performance with increasing batch size, while NRC~\cite{NRC}, AaD~\cite{AaD} and LLN~\cite{lln2023source} remain stable. This is because more samples result in more complex correlations, which highlights the effectiveness of high-order relations. To ensure a fair comparison, we use $64$ for all datasets.}

\begin{figure}[!t]
    \centering
    \begin{minipage}[t]{0.48\linewidth}
    \centering
    \includegraphics[width=0.48\linewidth]{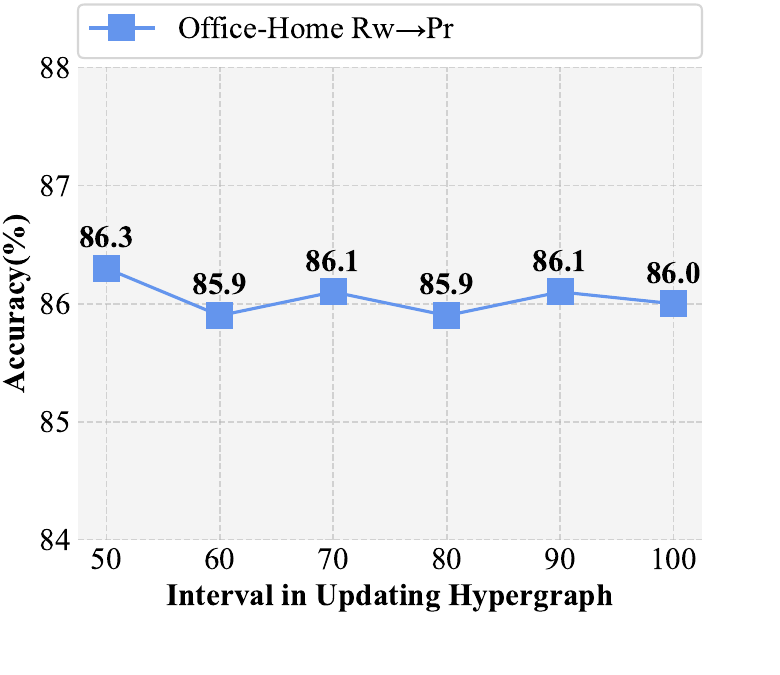}
    \includegraphics[width=0.48\linewidth]{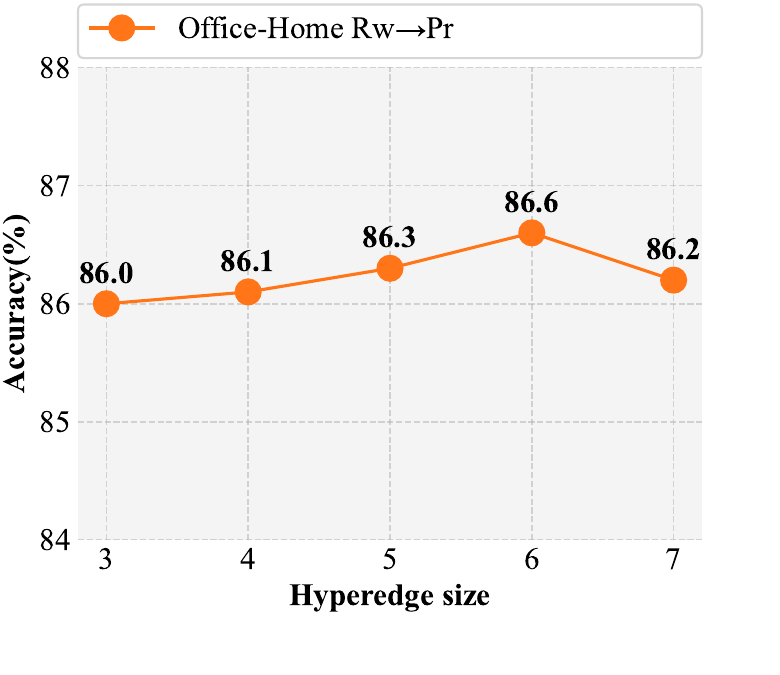}
    \caption{(Left) Effect of different intervals in updating hypergraph. (Right) Effect of different hyperedge degrees. }
    \label{fig:short3}
    \end{minipage}
    \hfill
    \begin{minipage}[t]{0.48\linewidth}
    \centering
    \includegraphics[width=0.48\linewidth]{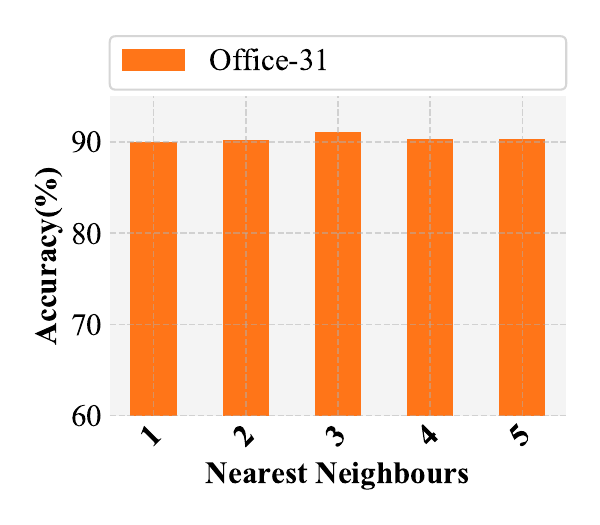}
    \includegraphics[width=0.48\linewidth]{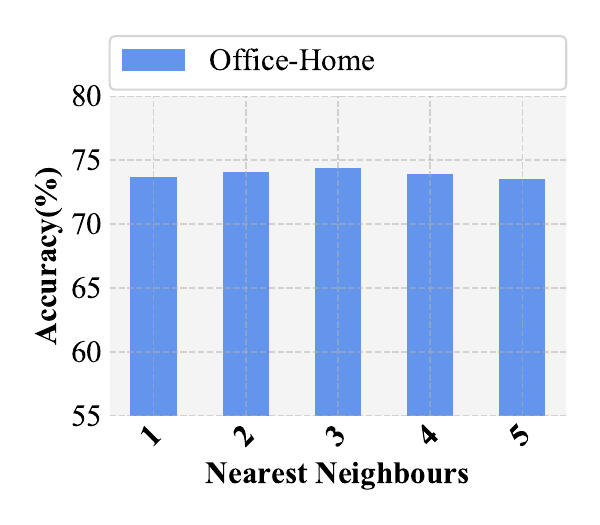}
    \caption{Effect of different numbers of nearest neighbors on Office-31 (Left) and Office-Home (Right).
    }
   \label{fig:short4}
    \end{minipage}
    \vspace{-0.5cm}
\end{figure}

\smallskip
\noindent \textbf{Interval $T_{in}$ in Updating Hypergraph.}
The constructed hypergraph is periodically updated during the training process to maintain its effectiveness. To investigate the effect of the update interval, we conducted an experiment on the Rw $\rightarrow$ Pr track on the Office-Home dataset. Fig.~\ref{fig:short3} (Left) shows the effect of using different update intervals of our method, showing that the accuracy of our method is stable at around $86\%$ when the interval number increases from $50$ to $100$. This indicates that the performance of our method is not sensitive to the update interval. In the main experiment, we select $50$ as the final interval number.

\smallskip
\noindent \textbf{Degree $k$ in Hyperedges. } Fig.~\ref{fig:short3} (Right) shows the effect of different degree $k$ in hyperedges. The experiment setting is the same as above. Specifically, we change the hyperedge degree k from $3$ to $7$ and find that our method is only slightly affected, which indicates that the hyperedge is also not sensitive to degree number. 

\smallskip
\noindent \textbf{Number of Nearest Neighbors in Clustering.} This part studies the effect of using different nearest neighbors in clustering. Fig.~\ref{fig:short4} shows the corresponding performance on Office-31 (Left) and Office-Home (Right) using different nearest neighbors $[1,5]$. From the figure we can observe that by using the median value $3$ achieves the best performance on these two datasets. This is due to that a small number of neighbors is prone to be affected by outliers and a mass of neighbors may lose the locality of group information.

\begin{figure}[t]
    \centering
    \begin{minipage}[b]{0.48\linewidth}
    \centering
    \includegraphics[width=0.48\linewidth]{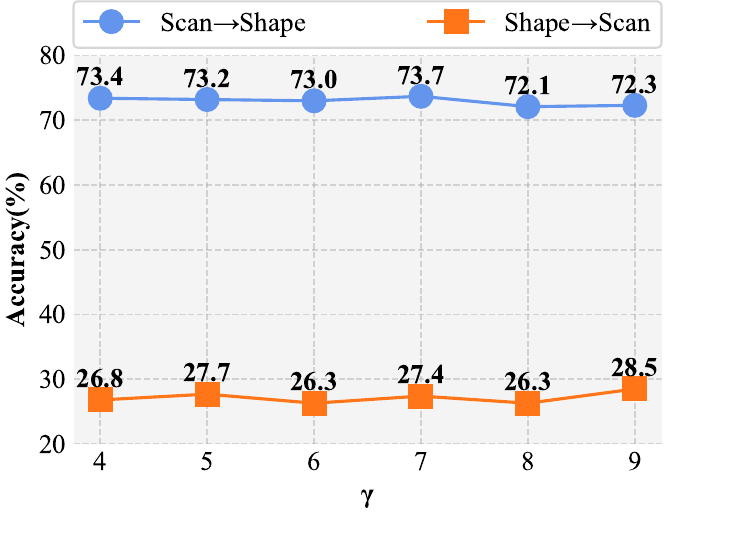}
    \includegraphics[width=0.48\linewidth]{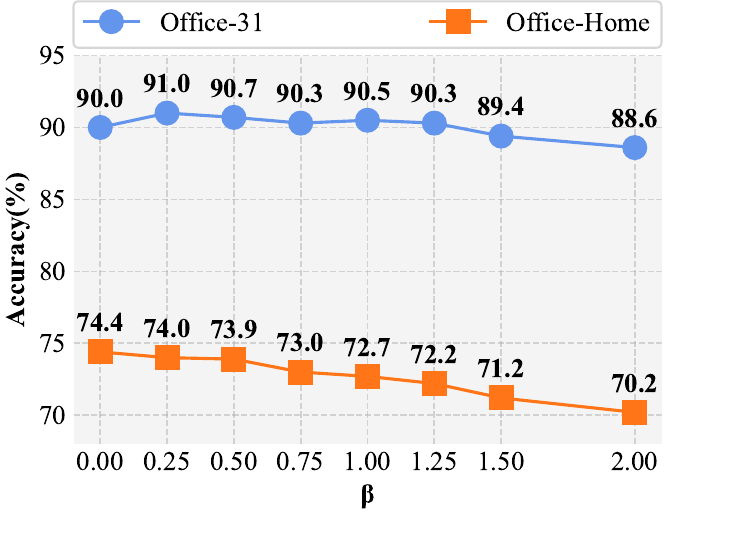}
    \caption{(Left) Effect of different scale factor $\gamma$. (Right) Effect of different balancing factor $\beta$.}
     \label{fig:short5}
    \end{minipage}
    \hfill
    \begin{minipage}[b]{0.5\linewidth}
    \centering  
   \begin{subfigure}{0.45\linewidth}
    \centering
     \includegraphics[width=\linewidth]{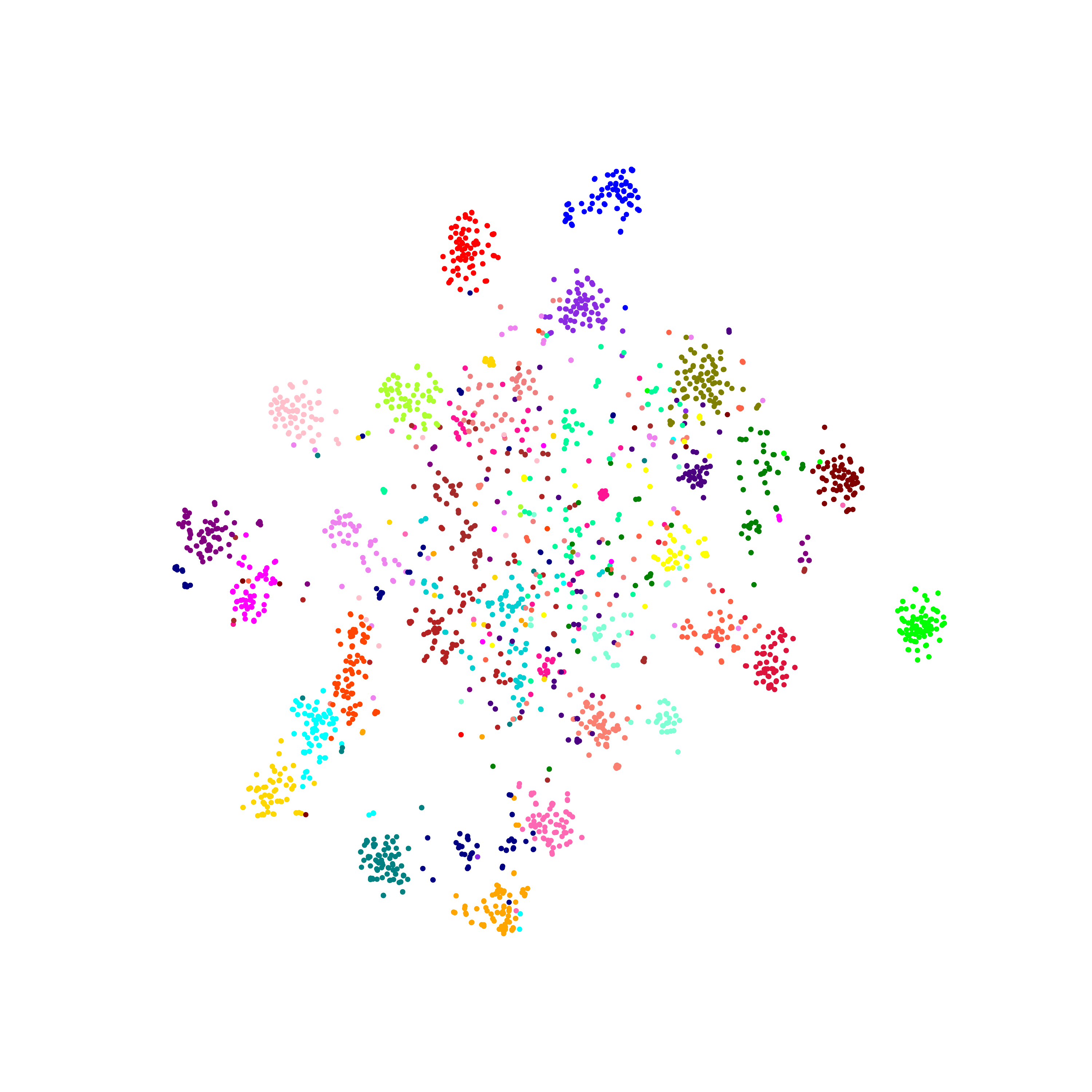}
     \caption{Before}
     \label{fig:short5-a}
   \end{subfigure}
   \hspace{0.05\linewidth} 
      \begin{subfigure}{0.45\linewidth}
     \centering
     \includegraphics[width=\linewidth]{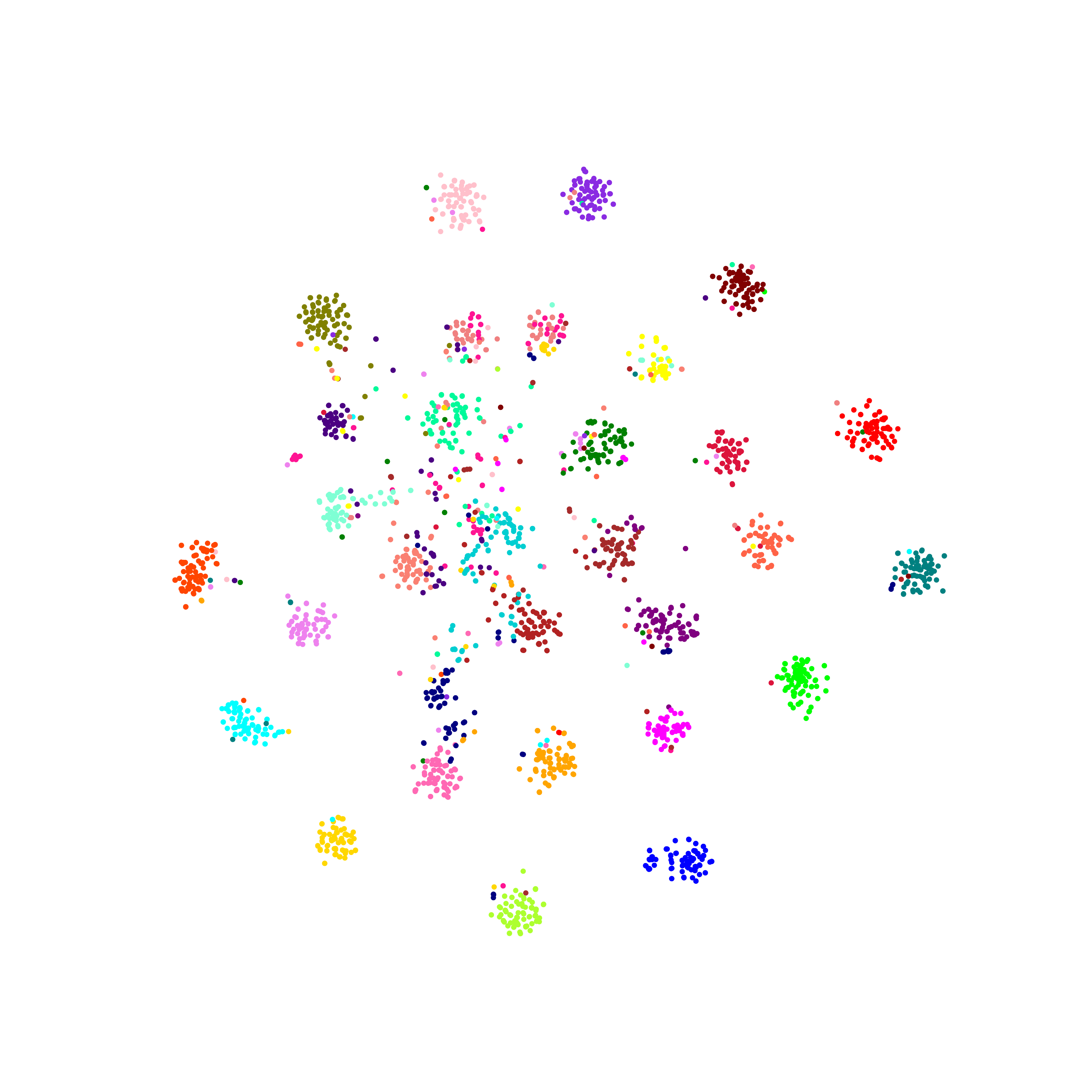}
     \caption{After}
     \label{fig:short5-b}
   \end{subfigure}
   \vspace{-0.2cm}
    \caption{T-SNE feature visualizations before and after domain adaptation.}
   \label{fig:short6}
    \end{minipage}
\end{figure}

\smallskip
\noindent \textbf{Scale factor $\gamma$ and hyperparameter $\lambda$ in Objective.} As shown in Eq.~\eqref{eq:important9}, $\gamma$ is the scale factor controlling the attention levels. Fig.~\ref{fig:short5} (Left) shows the effect of different $\gamma$ using two tracks of Scan $\rightarrow$ Shape and Shape $\rightarrow$ Scan on the PointDA-10 dataset, showing that our method is not sensitive to $\gamma$ either. We experimentally prove that taking ${\gamma=7}$ is valid for all datasets.
We then study the effect of different $\lambda$ on the second loss term. Since the rate of decay is controlled by a balancing factor $\beta$ as ${\lambda=(1+10 \cdot \frac{\text{iter}}{\text{maxiter}})^{-\beta}}$, we study the effect of $\beta$ instead.
Fig.~\ref{fig:short5} (Right) shows the effect of different $\beta$ on Office-31 and Office-Home datasets, which indicates that a proper $\beta$ is important and depends on the specific dataset distribution, \eg, $\beta = 0.25$ corresponds to the best on Office-31 and $\beta = 0$ for Office-Home dataset. 




\begin{figure}[t]
    \centering
    \begin{minipage}[b]{0.4\textwidth}
     \centering
     \footnotesize
     \captionof{table}{Source-free open-set Domain Adaptation on Office-Home.}
     
     \setlength{\tabcolsep}{1.2mm}{
     \scalebox{0.8}{
      \begin{tabular}{lccc}
       \hline
    Method & SF & Avg \\ \hline
    ResNet~\cite{he2016deepresnet50} & × & 65.3 \\
    OSBP (ECCV’18)~\cite{osbp} & × & 65.7 \\
    STA (CVPR’19)~\cite{STA} & × & 69.5 \\
    GLC (CVPR’23)~\cite{GLC} & × & 69.8 \\ \hline
    SHOT (ICML’20)~\cite{shot} & \checkmark & 72.8 \\
    SHOT-IM (ICML’20)~\cite{shot} & \checkmark & 71.5 \\
    SHOT+HCL (NeurIPS’21)~\cite{HCL} & \checkmark & 73.2 \\
    CoWA-JMDS (ICML’22)~\cite{CoWA-JMDS} & \checkmark & 73.2 \\
    AaD (NeurIPS’22)~\cite{AaD} & \checkmark & 71.8 \\
    U-SFAN (ECCV’22)~\cite{U-SFAN} & \checkmark & 73.5 \\
    CREL (CVPR’23)~\cite{zhang2023class} & \checkmark & 73.3 \\
    \rowcolor{mygray}Ours & \checkmark & \textbf{77.3} \\ \hline
    \end{tabular}}}
    \label{table_ip:7}
    \end{minipage}
    \hfill
    \begin{minipage}[b]{0.55\textwidth}
        \centering  
       \begin{subfigure}{0.35\linewidth}
        \centering
         \includegraphics[width=\linewidth]{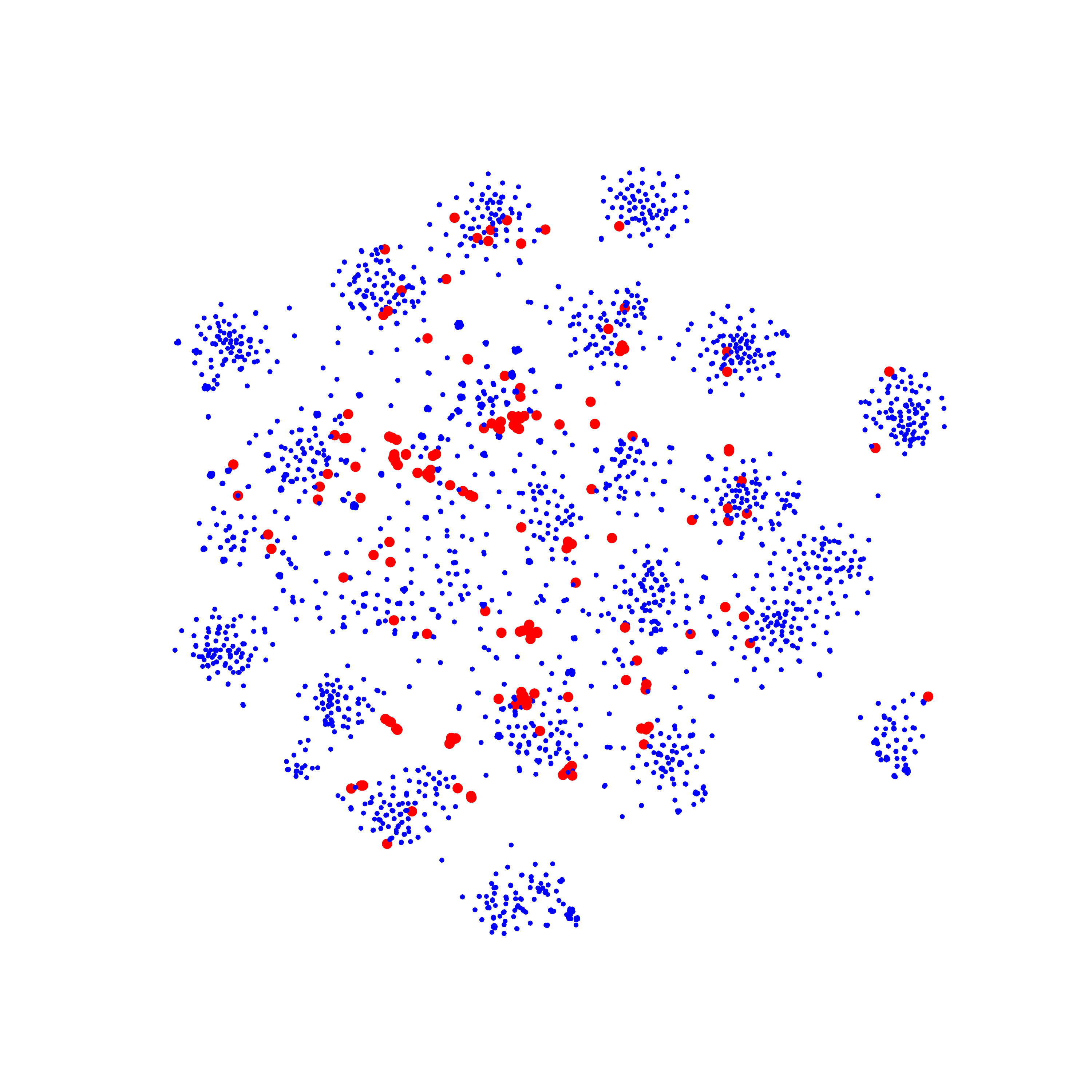}
         \caption{Before}
         \label{fig:short4-a}
       \end{subfigure}
       \hspace{0.05\linewidth} 
          \begin{subfigure}{0.35\linewidth}
         \centering
         \includegraphics[width=\linewidth]{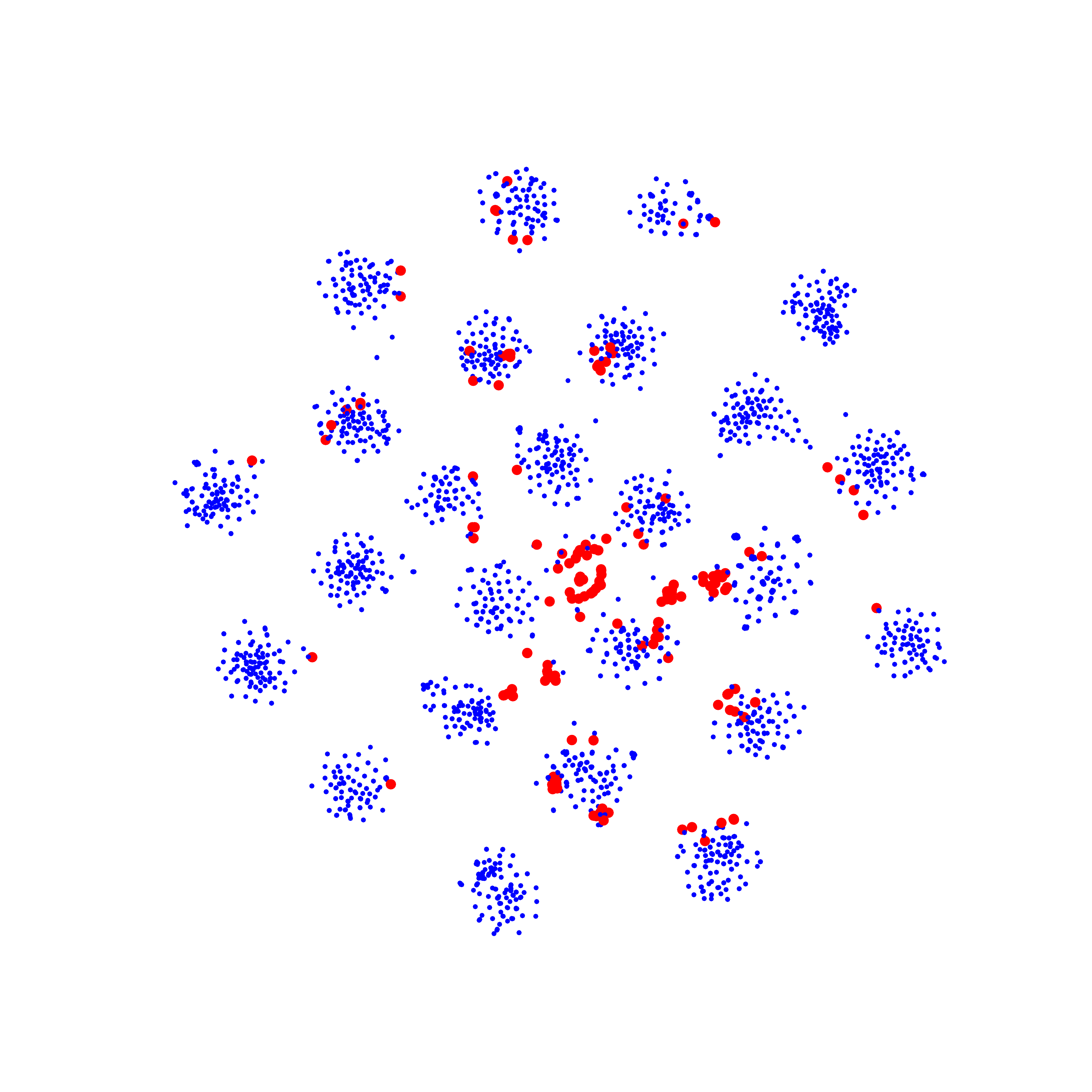}
         \caption{After}
         \label{fig:short4-b}
       \end{subfigure}
       \vspace{-0.2cm}
        \caption{T-SNE feature visualizations before and after adaptation. Blue and red colors correspond to known and unknown categories.}
       \label{fig:7}
    \end{minipage}
    \vspace{-0.5cm}
\end{figure}


\smallskip
\noindent \textbf{Feature Visualization.}
{Fig.~\ref{fig:short6} shows the t-SNE~\cite{van2008visualizing} visualization of features before adaptation and after adaptation. We can observe that after adaptation, the target samples of the same class become more coherent, and the margin between different classes becomes clearer and larger, demonstrating the effectiveness of our proposed method.}



\smallskip 
\noindent{\textbf{Source-free Open-set Domain Adaptation}
We provide additional experiments in the open-set DA setting on Office-Home. In the open-set scenario, the target domain includes unseen classes that are not contained in the source domain. For open-set detection, we follow the same protocol for the detection of unseen classes as in SHOT~\cite{shot}. We sort the entropy of the samples and perform two-class k-means clustering. The high entropy clusters are then classified as unknown samples and the low entropy clusters are classified as known samples. The known samples are used to train the model. As can be seen from the results in Table~\ref{table_ip:7}, our method outperforms the current state-of-the-art method~\cite{U-SFAN} with an improvement of $3.8\%$. {This scenario further highlights the benefit of high-order relations in uncovering the underlying correlations, especially the semantic difference between known and unknown categories, see Fig.~\ref{fig:7}.}


\vspace{-0.4cm}
\section{Conclusion}
\vspace{-0.2cm}
This paper introduces a new SFDA method called {\em Hyper-SFDA} that aims to exploit high-order neighborhood relations and explicitly take the domain shift effect into account. Specifically, we construct hyperedges over the target samples by considering their semantic similarity and develop a self-loop strategy to involve the domain uncertainty of target samples in hypergraph optimization. Then we propose an adaptive relation-based objective that pushes close samples in a cluster and pulls away samples in different clusters with soft attention levels. The experiments conducted on mainstream datasets have demonstrated the efficacy of our method on the SFDA problem.

%
%

\bibliographystyle{splncs04}
\bibliography{main}
\end{document}